\documentclass[letterpaper, 10 pt, conference]{ieeeconf}  
\IEEEoverridecommandlockouts                              

\makeatletter
\let\NAT@parse\undefined
\makeatother

\usepackage[english]{babel}             

\usepackage{amsmath}                    
\usepackage{amssymb}                    
\usepackage{mathtools}                  
\usepackage{dsfont}                     
\usepackage{bm}                         

\usepackage{graphicx}                   
\usepackage{animate}                    
\usepackage{epsfig}                     
\usepackage{epstopdf}                   
\usepackage{caption}                    
\usepackage{wrapfig}                    
\usepackage{subcaption}                 
\usepackage{booktabs}                   
\usepackage{multirow}                   
\usepackage{multicol}                   
\usepackage{makecell}                   
\usepackage{floatflt}                   
\usepackage{adjustbox}                  
\usepackage{lscape}                     

\usepackage{xcolor}                     
\usepackage{color}                      

\usepackage[colorlinks=true, linkcolor=blue, citecolor=green, urlcolor=blue]{hyperref}
\usepackage[square,numbers,sort&compress]{natbib} 
\usepackage{algorithm}                  
\usepackage{algorithmic}                
\usepackage{listings}                   

\usepackage{epigraph}                   
\usepackage{nomencl}                    
\usepackage{etoolbox}                   
\usepackage{siunitx}                    
\usepackage{textcomp,mathcomp}          
\usepackage{nameref}                    
\usepackage{url}                        
\usepackage{placeins}                   

\usepackage{multirow}
\usepackage{booktabs} 
\usepackage{arydshln} 

\newcommand{\link}[1]{\colora{\url{#1}}}

\renewcommand{\sec}[1]{Section~\ref{#1}}
\newcommand{\fig}[1]{Fig.~\ref{#1}}

\newcommand{\tab}[1]{Table~\ref{#1}}

\newcommand{\sensor}[0]{DTactive} %

\definecolor{lightorange}{HTML}{FFBE7A}
\definecolor{lightblue}{HTML}{5EAAF6}
\definecolor{lightred}{HTML}{FF5353}

\title{ \LARGE \bf \sensor: A Vision-Based Tactile Sensor with Active Surface}

\author{Jikai Xu$^{*,1,2}$, Lei Wu$^{*,1,2}$, Changyi Lin$^{3}$, Ding Zhao$^{3}$ and Huazhe Xu$^{1,4,5}$  \\
\url{https://ieqefcr.github.io/DTactive/}
\thanks{$^{*}$Jikai Xu and Lei Wu contribute equally.}
\thanks{Huazhe Xu is the corresponding author.}
\thanks{$^1$Shanghai Qi Zhi Institute, Shanghai, China}
\thanks{$^2$Huazhong University of Science and Technology}
\thanks{$^3$Carnegie Mellon University}
\thanks{$^4$Tsinghua University}
\thanks{$^5$Shanghai AI Laboratory}
}

\begin{document}
\maketitle

\begin{abstract}
	The development of vision-based tactile sensors has significantly enhanced robots' perception and manipulation capabilities, especially for tasks requiring contact-rich interactions with objects. In this work, we present \sensor, a novel vision-based tactile sensor with active surfaces. \sensor~inherits and modifies the tactile 3D shape reconstruction method of DTact while integrating a mechanical transmission mechanism that facilitates the mobility of its surface. Thanks to this design, the sensor is capable of simultaneously performing tactile perception and in-hand manipulation with surface movement. Leveraging the high-resolution tactile images from the sensor and the magnetic encoder data from the transmission mechanism, we propose a learning-based method to enable precise angular trajectory control during in-hand manipulation. In our experiments, we successfully achieved accurate rolling manipulation within the range of $\left[ -180^\circ,180^\circ \right]$ on various objects, with the root mean square error between the desired and actual angular trajectories being less than $12^\circ$ on nine trained objects and less than $19^\circ$ on three novel objects. The results demonstrate the potential of \sensor~for in-hand object manipulation in terms of effectiveness, robustness and precision.
\end{abstract}

\section{Introduction}

In-hand manipulation, a crucial skill for robots to achieve human-like dexterity, has garnered significant research attention in recent years. Beyond the need for dexterity, which ensures sufficient degrees of freedom for manipulating objects efficiently, another key challenge lies in the facts that visual occlusion is imminent and fine force control is necessary in contact-rich manipulation tasks. Therefore, dexterity and tactile sensing capability are two critical factors for manipulators to achieve efficient and robust in-hand manipulation.

The common design for manipulators capable of performing in-hand manipulation is the multi-articulated dexterous hand~\cite{shaw2023leap, wei2024wearable, si2024deltahands}. Augmented by tactile sensing, especially vision-based tactile sensors~\cite{yuan2017gelsight} that provide rich contact information, such manipulators can learn and perform complex manipulation skills~\cite{qi2023general, romero2024eyesight, ma2024gellink}. However, their high degree of freedom also leads to costly and complex control methods. Furthermore, scaling the sensing areas of vision-based tactile sensors for them presents a significant technical challenge.

Another line of work develops grippers with continuous activate surfaces~\cite{yuanDesignControlRoller2020,ko2020tendon,li2023active,cai2023hand,xiang2024adaptive,jiang2024rotipbot}, which also highlights a promising and effective approach for dexterous in-hand manipulation. To further enhance the robustness of manipulation, Tactile-Reactive Roller Grasper~\cite{yuan2023tactile} integrates tactile sensing into its rollers, but the cylindrical shape of the roller limits both its grasping capability and manipulation stability.
\begin{figure}[t]
	\centering
	\includegraphics[width= \linewidth]{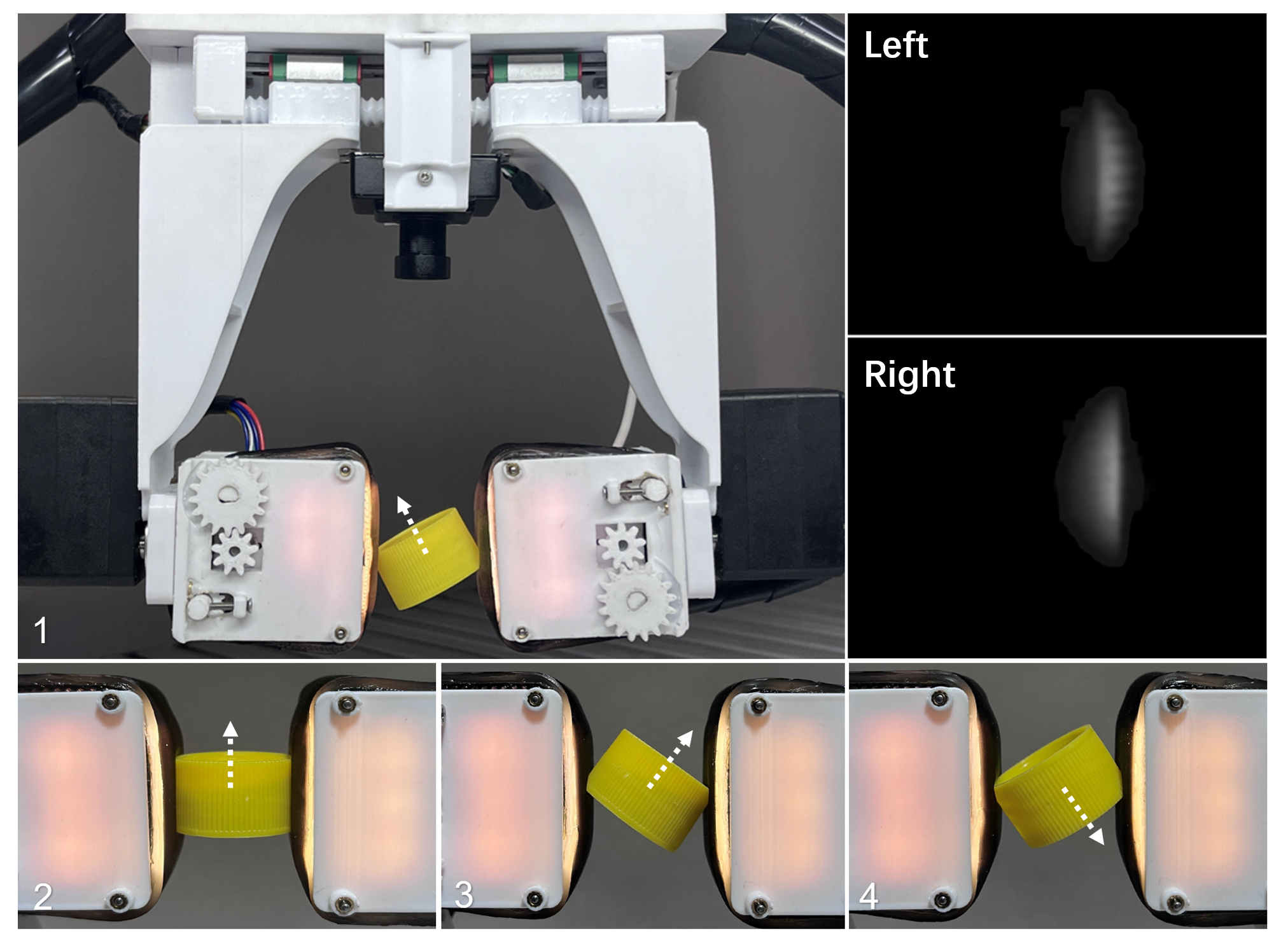}
	\caption{
		A two-finger gripper is equipped with \sensor~at its ends, using the tactile feedback and active surface of the sensors to perform bottle cap flipping manipulation. The numbered images represent the sequential steps of the operation.}
	\label{fig:headpic}
\end{figure}

In this work, we present \sensor, a vision-based tactile sensor with continuous active surfaces. \sensor~builds upon and adapts the tactile 3D shape reconstruction method presented in~\cite{lin2023dtact}, incorporating a mechanical transmission mechanism to enable the mobility of its surface. Additionally, its smaller corner radius, compared to a roller of the same size, enhances its versatility for use with various types of grippers. To explore the feasibility of \sensor~in object rolling manipulation, we mount two sensors at the end of a two-finger gripper as shown in \fig{fig:headpic}.
Furthermore, we develop a robust orientation estimator that enables precise rolling manipulation within the range of $\left[ -180^\circ,180^\circ \right]$ across various objects. The root mean square error of the tracking angle is less than $12^\circ$ for nine trained objects and less than $19^\circ$ on three novel objects, demonstrating the effectiveness, robustness, and precision of \sensor{}.

The remainder of this paper is organized as follows: we first introduce related work on hardware design and manipulation method in~\sec{sec:related}. The details of the design and fabrication of \sensor~are described in~\sec{sec:design}. We then present our control method for object rolling manipulation in ~\sec{sec:methods}. Next,~\sec{sec:results} reports the quantitative results of our experiments. Finally, the conclusion is summarized in~\sec{sec:conclusion}.

\section{Related Work}
\label{sec:related}

\subsection{Manipulators with Dexterity and Tactile Sensing}
Dexterous robotic hands with multiple articulated fingers~\cite{shaw2023leap, wei2024wearable, si2024deltahands} exhibit high dexterity through human teleoperation.
When covered with tactile films that provide sparsely distributed force information~\cite{yang2023tacgnn, yin2023rotating, guzey2023dexterity}, dexterous policies for limited skill sets can be effectively learned and deployed. To obtain more detailed tactile information of the contact object, the widely-used vision-based tactile sensor GelSight~\cite{yuan2017gelsight}, which can reconstruct high-resolution 3D contact geometry, has been integrated into dexterous hands~\cite{qi2023general, romero2024eyesight, ma2024gellink}. However, fully equipping the fingers and palm with tactile sensing remains a challenge due to the difficulty of scaling the sensing areas of vision-based tactile sensors within a compact form factor~\cite{zhao2023gelsight, van2020large}.

Compared to dexterous hands, roller grippers featuring continuous active surfaces~\cite{tincani2012velvet, yuan2020design, cai2023hand} offer comparable or even superior dexterity and efficiency in certain tasks such as object rolling. Building on the design of roller grippers for advanced dexterity and the optical principles of GelSight for tactile sensing, the Tactile-Reactive Roller Gripper (TRRG)~\cite{yuan2023tactile} achieves efficient continuous rolling of unknown objects~\cite{lepert2023hand}. However, TRRG has the following notable drawbacks.
First, the key photometric stereo technique~\cite{johnson2009retrographic} used in GelSight requires uniform colorful lights from different directions parallel to the surface. This requirement is extremely difficult to meet for a continuous active surface moving along the surface direction. Consequently, TRRG faces complex design and fabrication challenges and compromises the 3D shape reconstruction capability when using the same LED configurations as in~\cite{tippur2023gelsight360}.
Furthermore, the cylindrical surface of the TRRG significantly limits its dexterity. On one hand, this design creates gaps beneath the contact area of the two rollers, making it unable to grasp small objects on a flat surface. On the other hand, the cylindrical shape reduces the tactile contact area, resulting in low stability during object manipulation.

Instead of using the photometric stereo technique, \sensor{} inherits the optical principles of DTact-type tactile sensors~\cite{lin2023dtact, lin20239dtact} for 3D shape reconstruction. This method demonstrates comparable accuracy, as well as robustness and surface shape extensibility, while requiring only perpendicular white illumination to the surface. This approach allows \sensor{} to maintain accurate 3D shape reconstruction with minimal added complexity in design and fabrication.
Furthermore, with an optimized transmission mechanism and proposed fabrication processes, \sensor{} achieves a square-shaped surface, significantly enhancing stability and adaptability to objects of various sizes.



\subsection{Tactile-Based Orientation Estimation}
There are two typical categories of model-based tactile-based object pose estimation methods: (i) marker-based methods~\cite{ma2021extrinsic, liu2023enhancing}, which involve the use of markers printed on the tactile sensor's surface to track the motion of contact points between the object and the sensor, and (ii) point-cloud-based methods~\cite{lin2023dtact, wang2021gelsight}, which utilize point cloud registration techniques, such as Iterative Closest Point (ICP) method~\cite{rusinkiewicz2001efficient}, for pose estimation. In the former category, marker points affect the quality of tactile reconstruction and increase the complexity of sensor fabrication, whereas in the latter category, point-cloud-based methods may encounter degradation issues when handling objects with shapes such as spheres and cylinders.
Instead of relaying only tactile images, our estimation method utilizes both the tactile images and the odemotry information from the magnetic encoders, which achieves robust estimation on various objects.

\section{Design, Fabrication, and Dexterity Analysis} 
\label{sec:design}
	\begin{figure}[t]
	\centering
	\includegraphics[width= \linewidth]{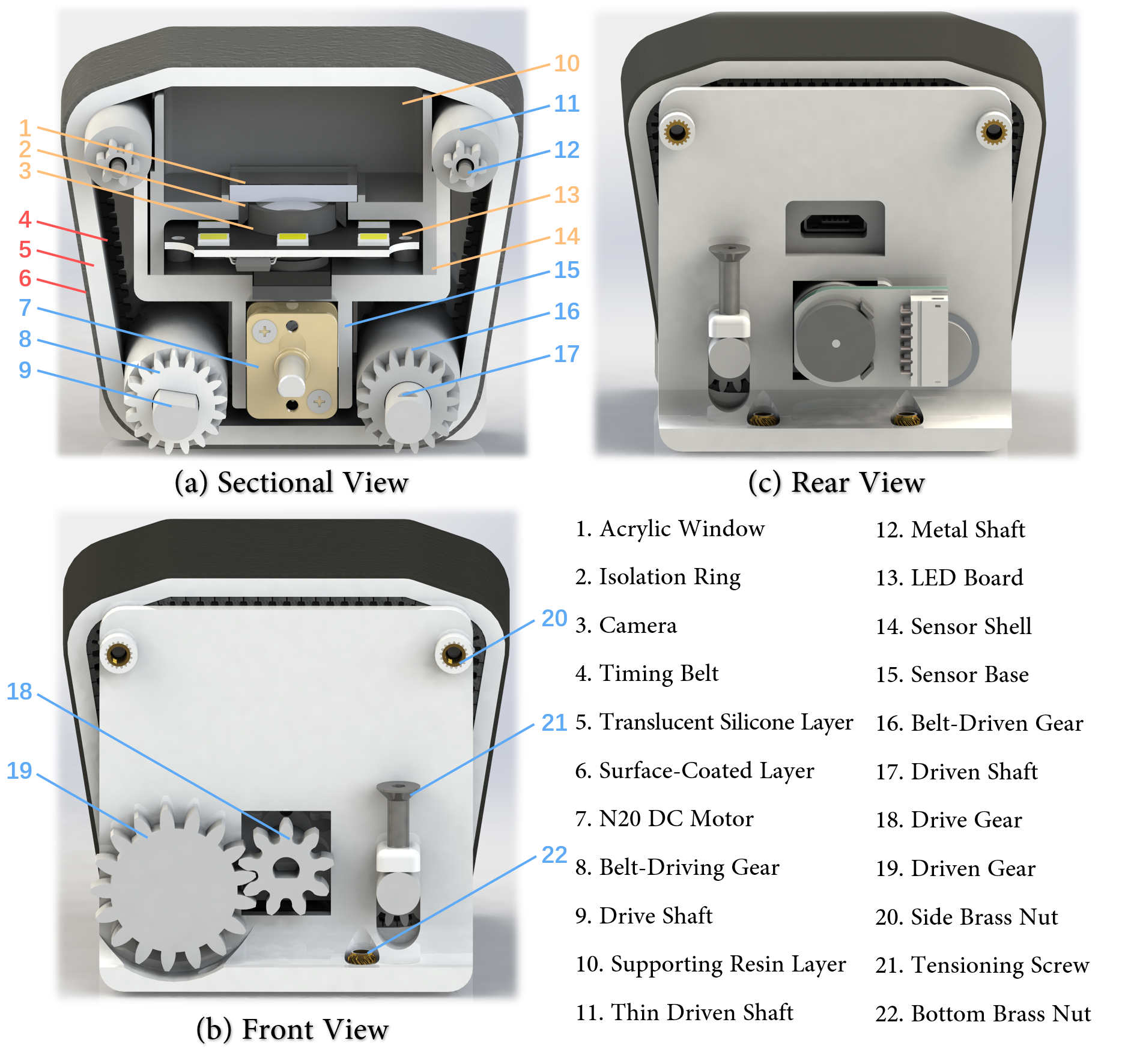}
	\caption{Design of \sensor{}. (a) Components of \sensor{} are shown in the sectional view. (b) Front view of~\sensor, showing the external transmission mechanism. (c) Rear view of ~\sensor.}
	\label{fig:design}
\end{figure}


 

In this section, our objective is to introduce the design and fabrication process of \sensor{}, and theoretically analyze its dexterity. To integrate the tactile sensing system of DTact~\cite{lin2023dtact} into \sensor{} , many improvements are made in terms of material, structure design and fabrication process. In addition, we analyse \sensor{}'s dexterity including the minimum size of objects being grabbed, the anti-torsion stability, and the maximum lifting force.

\subsection{Design and Fabrication}
As illustrated in~\fig{fig:design} (a), there are mainly three sub-modules of \sensor{}, including the optical sensing module (with parts in \textcolor{lightorange}{orange color}), the contact module (with parts in \textcolor{lightred}{red color}) and the mechanical module (with parts in \textcolor{lightblue}{blue color}).

\begin{figure}[t]
	\centering
	\includegraphics[width= \linewidth]{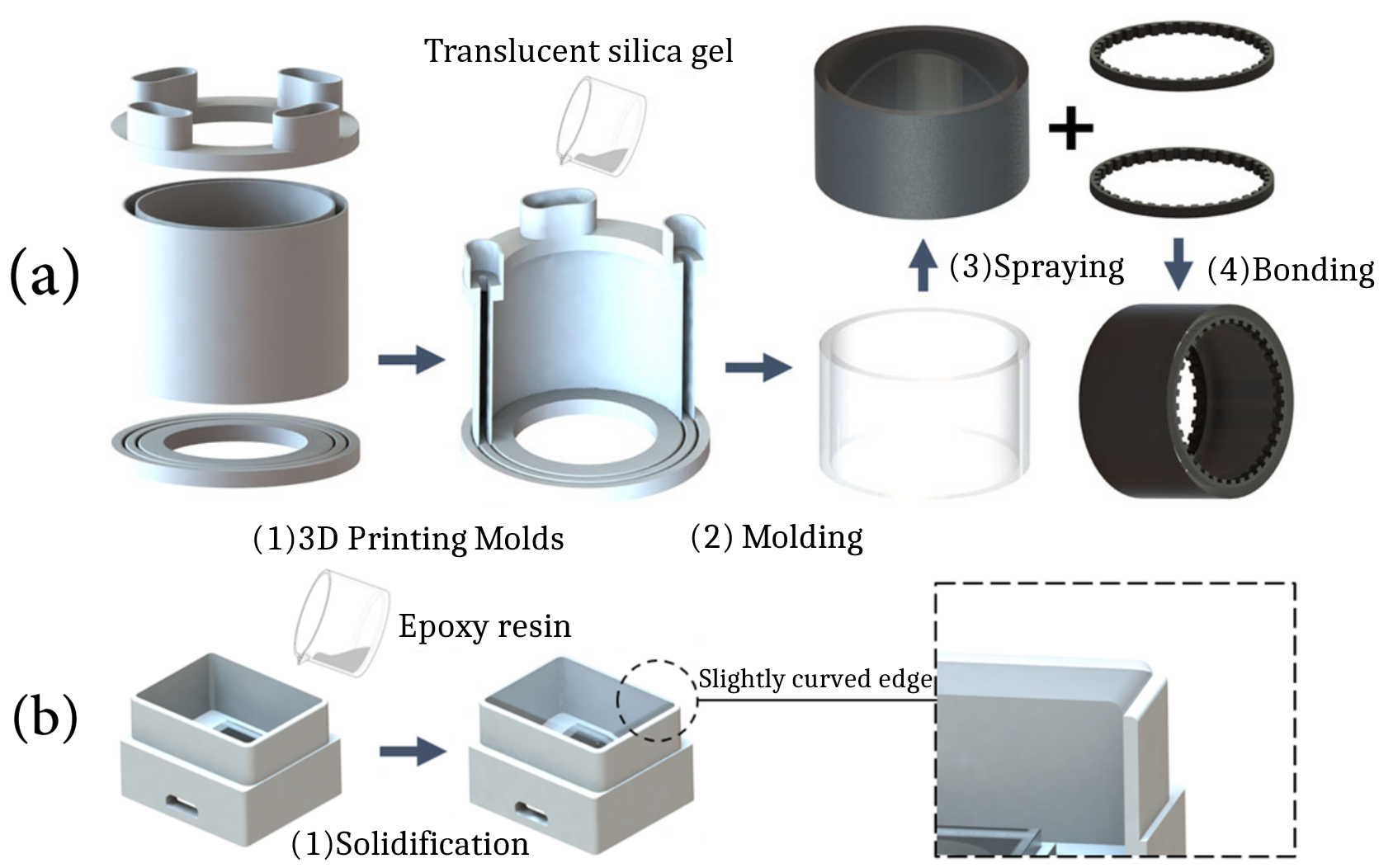}
	\caption{Fabrication of the essential components of \sensor{}. (a) Fabrication of the contact module. (b) Fabrication of the supporting resin layer.}
	\label{fig:fabrication}
\end{figure}

\textbf{The optical sensing module.} The optical sensing module is responsible for managing light and capturing tactile image. A key component of this module is the supporting resin layer as shown in  \fig{fig:design} (a) item 10, which serves two primary functions in \sensor: diffusing the light from the LED board to achieve uniform illumination on the translucent silicone layer and providing structural support to maintain close contact to translucent silicone layer. Notably, the supporting resin layer is designed to possess slightly curved edges as shown in \fig{fig:fabrication} (b), which facilitates smooth sliding of the contact module.

\textbf{The contact module.} The contact module directly interacts with objects and generates tactile deformations that captured by the optical sensing module. It consists of the surface-coated layer, the translucent silicone layer and two timing belts. As shown in \fig{fig:fabrication} (a), we first use 3D printer to fabricate the casting molds. Next, the translucent silica gel is poured into the molds to form the silicone belt. After approximately 6 hours, we disassemble the molds and spray black silicone on the outer surface of the silicone belt. Subsequently, two annular timing belts are bonded to the edges of the inner surface of the silicone belt using silicone-specific adhesive.

\begin{figure}[t]
	\centering
	\includegraphics[width= \linewidth]{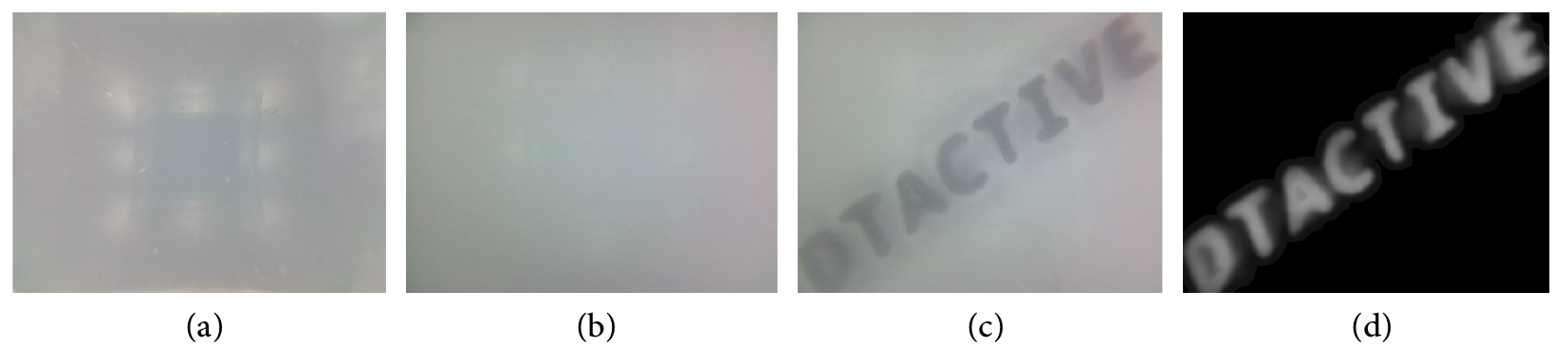}
	\caption{Tactile 3D reconstruction. (a) The tactile image without lubricating oil. (b) The tactile image with lubricating oil applied. (c) The tactile image when a badge with printed letters is pressed onto the sensor. (d) Depth map corresponding to the captured image in (c).}
	\label{fig:reconstruction}
\end{figure}

To enhance the quality of tactile images and ensure smooth sliding of the contact module, we apply lubricating oil between the translucent silicone layer and the supporting resin layer. As shown in the comparison between \fig{fig:reconstruction} (a) and (b), filling the lubricating oil effectively eliminates the strong reflections caused by the difference in refractive indices between resin and silicone. Consequently, \sensor{} achieves high-quality tactile 3D reconstruction, as illustrated in \fig{fig:reconstruction} (c) and (d).

\textbf{The mechanical module.}
The mechanical module encompasses both the structure and transmission mechanisms of the sensor. N20 DC motor is chosen as the actuation source for the active surface due to its compact size and ability to provide substantial torque. As shown in~\fig{fig:design} (a) and (b), the torque is transmitted to the drive shaft through the drive gear and the driven gear, and then the belt-driving gear drives the timing belt to convey the contact module. The belt-driven gear and two thin driven shafts are installed at the corner of the sensor to further reduce resistance. 


\subsection{Dexterity Analysis}
The established benchmark for \sensor{} is Tactile-Reactive Roller Grasper~\cite{yuan2023tactile}. 
We will evaluate their capability for picking small objects and stability of grasping. 

\textbf{Minimum radius that can pick up an object.} 
The sensor's ability to grasp objects depends on the radius of rounded corners of the fully contracted gripper, which limits the minimum radius of objects it can handle.

\begin{figure}
    \centering
    \includegraphics[width=1\linewidth]{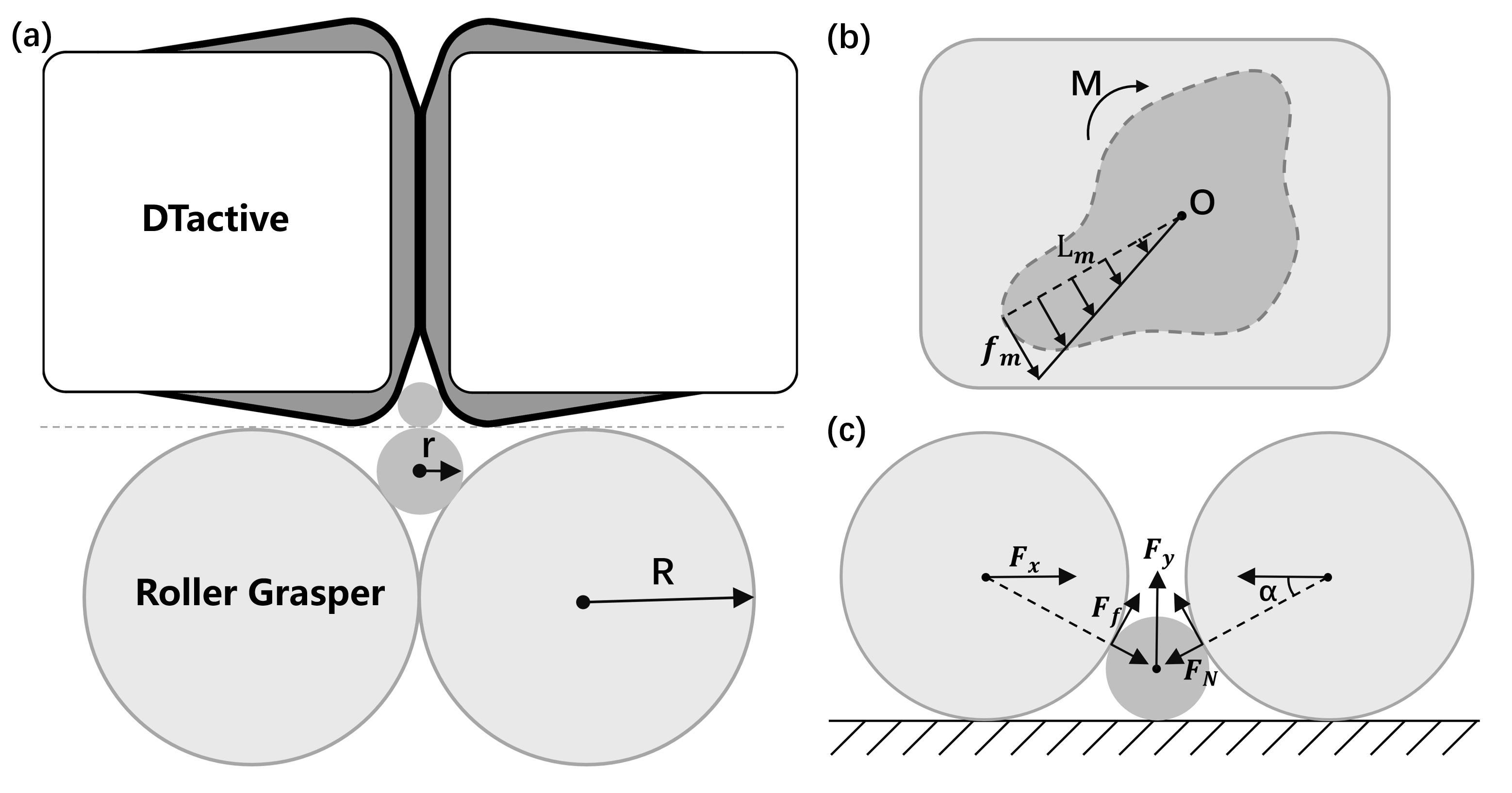}
    \caption{(a) Comparison of the minimum object radius that can be grasped by \sensor{} and a roller grasper. (b) Force analysis of an object under external moment (M) during grasping. (c) Force components involved in lifting an object. } 
    \label{fig:compare}
\end{figure}

As illustrated in \fig{fig:compare} (a), the minimum object radius $r$ that the roller grasper can handle is related to the radius of the gripper’s corner $R$, following the relationship $r = R/4$.
The radius of Roller Grasper is $20\text{mm}$, so the minimum radius of the object is $5\text{mm}$. For \sensor{}, Due to the protrusion of the sensor's tactile plane, its corners are not perfectly regular. A geometric analysis based on its basic dimensions indicates that the minimum radius of graspable objects is $2.81\text{mm}$.

\begin{figure*}[t]
	\centering
	\includegraphics[width=\linewidth]{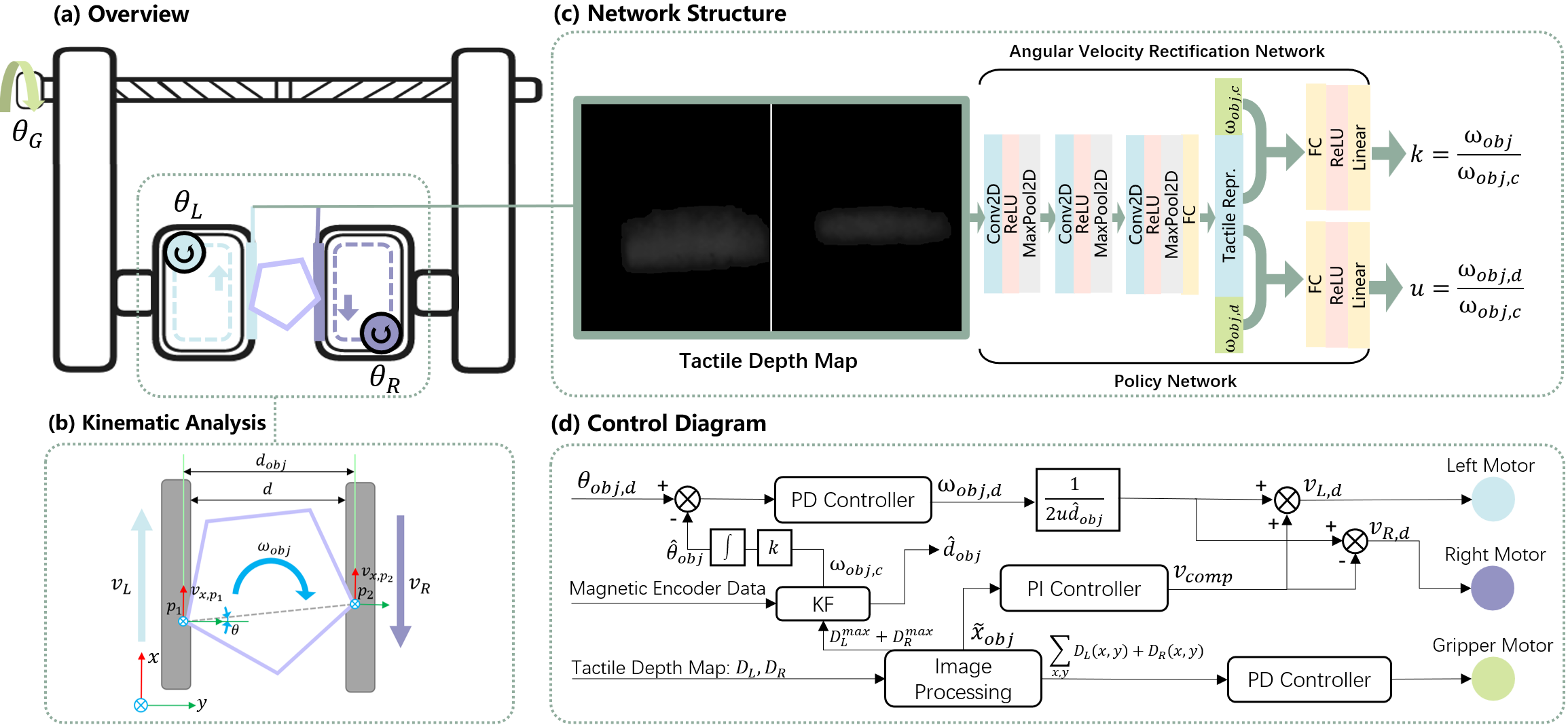}
	\caption{(a) A two-finger gripper equipped with two~\sensor{}.  (b) Kinematic analysis of the object manipulated by the gripper. (c) The structure of the angular velocity rectification network for orientation estimation and the policy network for generating angular velocity command. (d) Control diagram of the gripper.}
	\label{fig:pipline}
\end{figure*}

\textbf{Anti-torsion ability.} \sensor{} features an expanded tactile plane that increases the contact area with the object, thus improving the resistance to torsion. 
As illustrated in \fig{fig:compare}(b), the rectangle represents the tactile surface of \sensor{}, while the darker irregular shape denotes the contact area between the object and \sensor{}. Assuming the point $ O $ as the center of rotation, $ L_m $ represents the longest radial distance from the point $ O $ to the boundary. $ M $ denotes the external torque applied to the object and the contact area is $A_0$. The infinitesimal displacement of the mass points on the contact surface is directly proportional to their distance from the center $O$. Under small deformation conditions, silicone is considered to obey Hooke's law, which implies that the friction force is proportional to the displacement. Consequently, sliding will first occur at the point furthest from the center, where the maximum friction force is given by:
$$
df_{max} = \mu F_N = \mu\frac{F_{N0}}{A_0}dA
$$
where $F_{N0}$ is the total force exerted by the gripper on the object.
The force at a distance $l$ from the center $ df(l) = \frac{l}{L_m}df_{max} $, the sum of the torque provided by two sensors is then:
$$
T = 2\iint\limits_{A_0} l \, df(l) =\frac{2\mu F_{N0}I_p}{L_m A_0} 
$$
where $I_p$ represents the polar moment of inertia of the contact surface. This calculation confirms that the flat tactile surface provided by the \sensor{} significantly enhances the stability of grasping the object. Specifically, the expanded and more uniform contact area increases the polar moment of inertia $I_p$, as the contact area is expended while contact points are distributed more evenly and farther from the center of rotation. 


\textbf{Lifting force can be provided.}
Refer to \fig{fig:compare} (c), the equation for the lifting force $F_y$ can be derived as follows:
$$ F_y = 2(F_f\cos{\alpha} - F_N\sin{\alpha}) $$
where $F_f$ represents the frictional force, and the angle $ \alpha = \arctan{\left(\frac{R}{r} - 1\right)} $. As the radius $ R $ decreases, $ \alpha$ also decreases, leading to an increase in the frictional force $F_f$.
Therefore, when grasping the same object, \sensor{}, which possesses a smaller corner radius, can provide more friction force.


\section{Control for Object Rolling Manipulation}
\label{sec:methods}
	In this section, we will first describe the kinematic analysis of the object during rolling manipulation, followed by our learning-based object orientation estimation method. We then introduce the closed-loop control strategy used for rolling manipulation. The final subsection provides a detailed description of the data collection process.

\subsection{Kinematic Analysis}

As shown in \fig{fig:pipline} (a), a two-finger gripper with with two DTactive mounted is driven by three motors: one for the opening and closing of the gripper and two for driving the sensors. The angles of these motors, denoted as \(\theta_{G}\), \(\theta_{L}\), and \(\theta_{R}\) are obtained using magnetic encoders.

In \fig{fig:pipline} (b), We define the x-direction as parallel to the sensor surface, while the y-direction is perpendicular to it. \( p_1 \) and \( p_2 \) represent the points on the object that are pressed deepest into the left and right sensors, respectively. In cases where there are multiple points at the maximum depth, any one of them may be selected. According to geometric and kinematic relationships, we have:
\begin{equation}
\label{eq:geometric}
d_{obj} =  \|p_1 - p_2\| \cos{\theta} 
\end{equation}
\begin{equation}
\label{eq:kinematic}
v_{x,p_1} - \omega_{obj} \|p_1 - p_2\| \cos{\theta} = v_{x,p_2}
\end{equation}
where $d_{obj}$ represents the distance between $p_1$ and $p_2$ in the $y$ direction; \(\theta\) is the angle between the line connecting \( p_1 \) and \( p_2 \) and the $y$ direction; \( v_x\) represents the velocity component of a point along the $x$ direction; $\omega_{obj}$ represents the angular velocity of the object. Substituting \eqref{eq:geometric} into \eqref{eq:kinematic}, we obtain:
\begin{equation}
\label{eq:angular_velocity}
\omega_{obj} = \frac{1}{d_{obj}}(v_{x,p_1} - v_{x,p_2})
\end{equation}
Therefore, the angular velocity of the object can be calculated only using $v_{x,p_1}$, $v_{x,p_2}$ and $d_{obj}$.

\subsection{Orientation Estimator}

We first define the angular velocity command \( \omega_{obj,c} \) using the following equation:
\begin{equation}
\label{eq:rough_estimation}
\omega_{obj,c}  = \frac{1}{d_{obj}}(v_L + v_R)
\end{equation}
where \(v_L\) and \(v_R\) represent the surface velocity of the left and the right sensors, respectively. Due to the tangential deformation of the sensor surface during the rolling manipulation, there will be discrepancies between \(v_{x,p_1}\) and \(v_L\), as well as between \(v_{x,p_2}\) and \(v_R\).
We denote the ratio of the true angular velocity to the angular velocity command as $k$:
\begin{equation}
k = \frac{\omega_{obj}}{\omega_{obj,c}} = \frac{v_{x,p_1} - v_{x,p_2}}{v_L + v_R}
\end{equation}
In our preliminary experiments, $k$ was found to be highly related to tactile image and the angular velocity command. The value of \(k\) reflects the ease of rolling an object. For example, circular objects are easy to roll, with a \(k\) value close to 1. In contrast, square object are difficult to roll when its flat surface is in contact with the sensor, but they rotate more easily when its edges are in contact. Thus, we predict $k$ by training a neural network:
\begin{equation}
k = \mathcal{N}(D_L, D_R, \omega_{obj,c})
\end{equation}
where $D_L, D_R \in \mathbb{R}^{460 \times 345}$ represent the left and right tactile depth images respectively. The structure of the angular velocity rectification network $\mathcal{N}$ is shown in \fig{fig:pipline} (c). The angular velocity command can be derived from the magnetic encoder data of the three motors, multiplied by \(k\) to obtain a more accurate angular velocity estimate, and then integrated over time to estimate the object's orientation.

\begin{figure}[t]
	\centering
	\includegraphics[width=\linewidth]{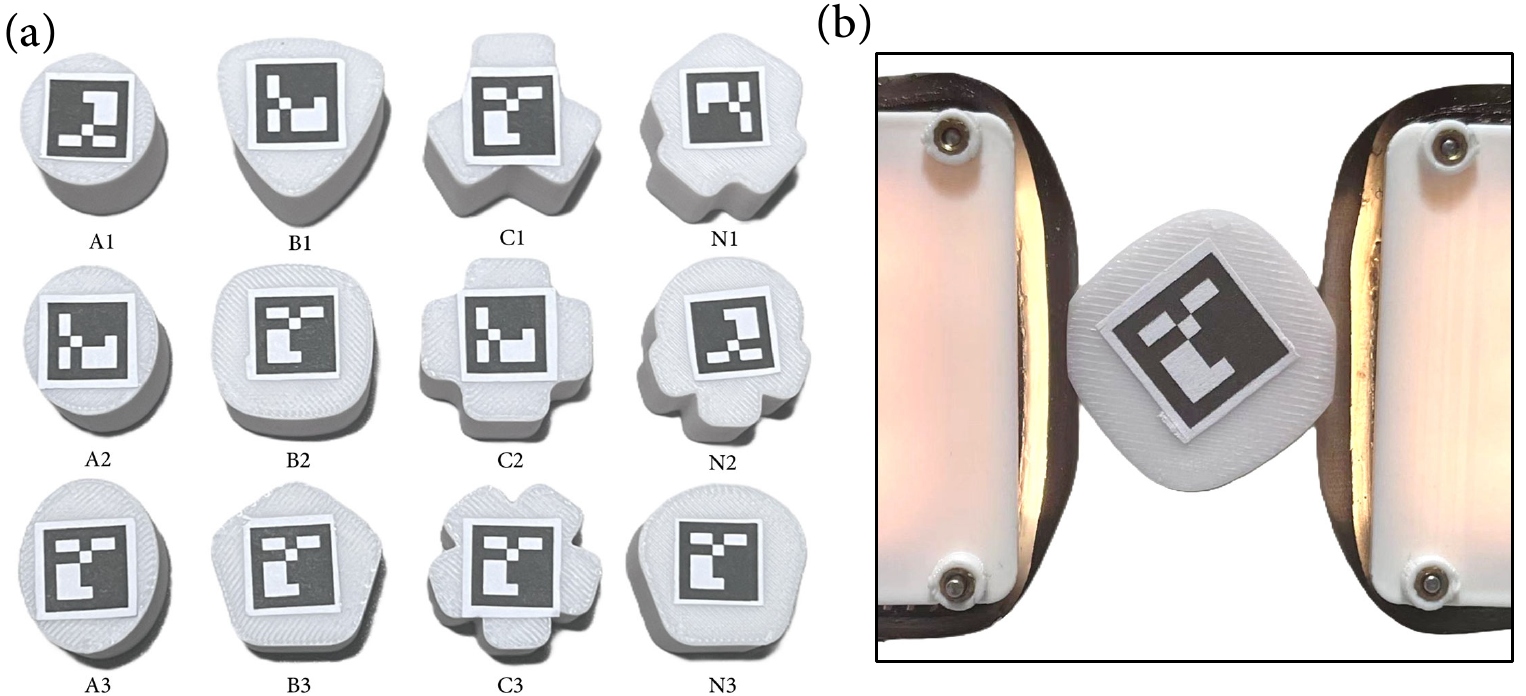}
	\caption{(a) 3D printed objects for rolling manipulation. A, B, C, and N represent three categories of objects (A, B, C) and novel objects (N), with the numbers indicating their index. (b) The gripper performs rolling manipulation of the object around the Z-axis of the ArUco Marker.}
	\label{fig:object}
\end{figure}

\begin{figure}[t]
	\centering
	\includegraphics[width=\linewidth]{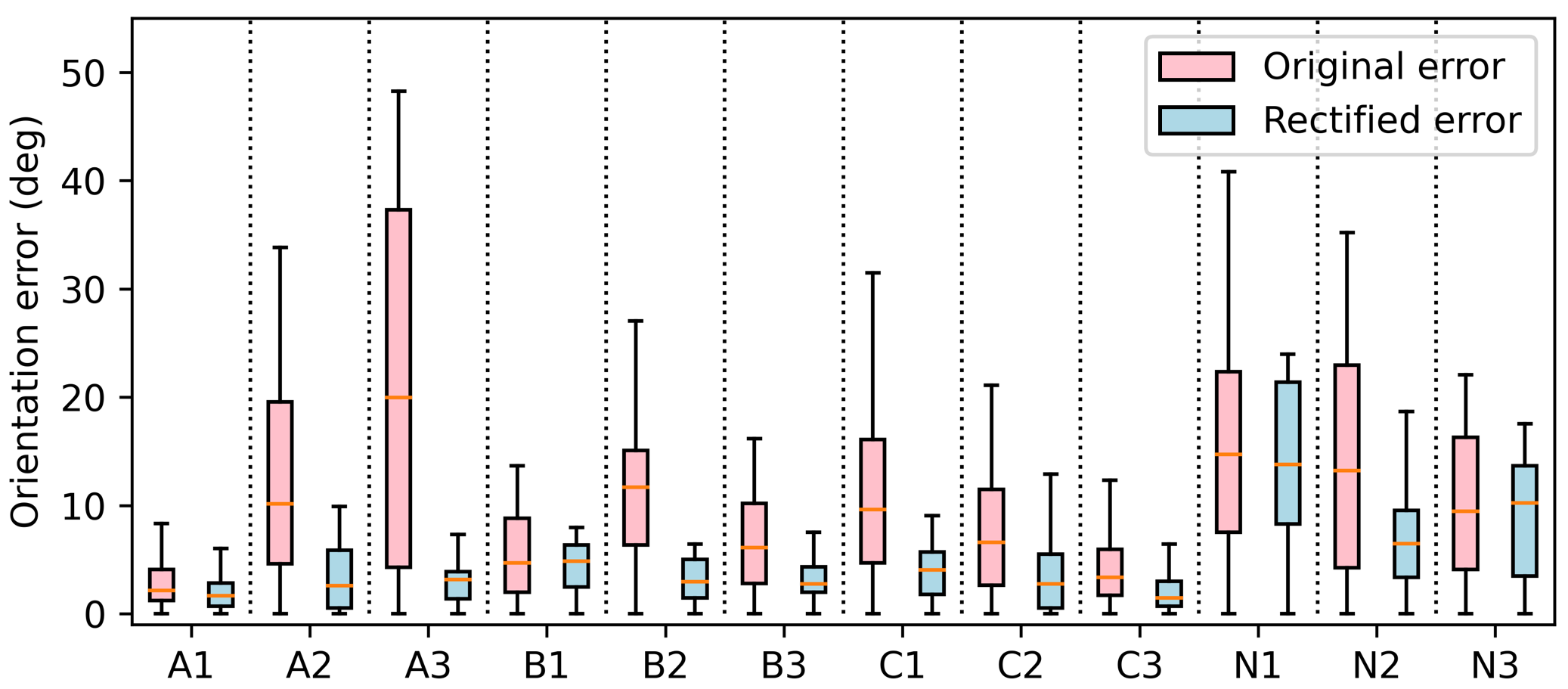}
	\caption{Orientation error of offline testing results.}
	\label{fig:offline_result}
\end{figure}

\subsection{Control Strategy}
To achieve robust object rolling manipulation, the gripper must rely on tactile feedback to control the three motors, implementing control in the following three aspects:

\textbf{Gripping force control.} Excessive or insufficient gripping force cannot achieve robust rolling manipulation, whereas tactile feedback can provide guidance for the appropriate gripping force. We use the sum of the tactile depth map $\sum_{x,y} D_L(x,y) + D_R(x,y)$ as the control metric for the gripping force. As shown in \fig{fig:pipline} (d), a PD controller is employed to maintain this sum at a specified appropriate value, thereby achieving proper control of the gripping force.

\textbf{Orientation control.} To achieve the desired object orientation $\theta_{obj,d}$, the error between $\theta_{obj,d}$ and the estimated orientation $\hat{\theta}_{obj}$ is processed through a PD controller to output the desired angular velocity $\omega_{obj,d}$. In contrast to the angular velocity rectification network, we train a policy network as shown in \fig{fig:pipline} (c) to predict the angular velocity command required to achieve the desired angular velocity:
\begin{equation}
    \omega_{obj,c} = \frac{\omega_{obj,d}}{u}\\
\end{equation}
\begin{equation}
    u = \pi(D_L, D_R, {\omega}_{obj,d})
\end{equation}

\begin{figure*}[ht]
	\centering
	\includegraphics[width=\linewidth]{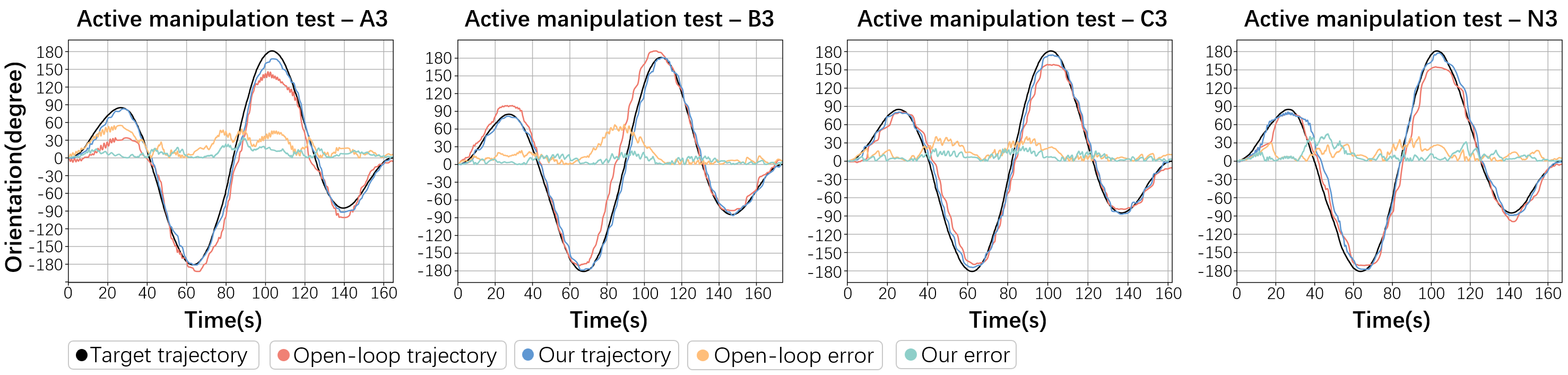}
	\caption{From left to right are the online testing trajectories of A3, B3, C3, and N3 under open-loop and our method.
}
	\label{fig:online_result}
\end{figure*}

\textbf{In-hand position control.} During object manipulation, the object should be kept as close to the center of the sensor as possible to prevent it from slipping out of the gripper. However, when the shape of the object is unknown and its initial position is random, it becomes barely feasible to precisely estimate and control the object's centroid position. Therefore, we use tactile images to obtain a rough estimate of the object's position along the sensor surface direction, denoted as $\tilde{x}_{obj}$, which is calculated as the average of the centroids of the left and right tactile images. As shown in \fig{fig:pipline} (d), the error between the center position of the sensor and \(\tilde{x}_{obj}\) is processed through a PD controller to output a compensation velocity \(v_{comp}\). Then, the desired velocities for the left and right surfaces are obtained using the following equations:
\begin{equation}
\left\{
\begin{aligned}
v_{L,d} &= \frac{\omega_{obj,c}}{2d_{obj}} + v_{comp} \\
v_{R,d} &= \frac{\omega_{obj,c}}{2d_{obj}} - v_{comp}
\end{aligned}
\right.
 \end{equation}
Consequently, while keeping \(\omega_{obj,c}\) unchanged, the position of the object within the hand can be adjusted.

\subsection{Data Collection}
We selected 12 types of objects as shown in \fig{fig:object} (a) to train and test our network. As illustrated in \fig{fig:object} (b), the gripper performs rolling manipulation on the object and obtains the ground truth of its angular velocity through the ArUco Marker~\cite{garrido2014automatic}. The object is manipulated by the gripper from 0 to 360 degrees, while trajectory data is recorded at a rate of 20 frames per second. For the nine objects in categories A, B, and C as shown in  \fig{fig:object} (a), we collected four trajectories under different angular velocity command for each object, resulting in a total of 36 trajectories and over 35000+ frames of data. For each novel object ($N1 \rightarrow N3$), we collected one trajectory to test the generalization performance of our model. We provide a more detailed description of the training and testing phases below.

\begin{itemize}
    \item \textit{Training:} We randomly select three out of the four trajectories for objects in categories A, B, and C to simultaneously train the angular velocity rectification network and the policy network. It is worth noting that the labels for both networks are \(\frac{\omega_{obj}}{\omega_{obj,c}}\). However, the former uses \(\omega_{obj,c}\) as input, whereas the latter employs \(\omega_{obj}\).
    
    \item \textit{Testing:} Testing consists of offline testing and online testing. Offline testing evaluates the accuracy of the orientation estimation by the angular velocity rectification model on trajectories that is not used for training. Online testing evaluates the performance of our method in controlling the object's orientation in accordance with a desired sinusoidal trajectory.
\end{itemize}

\section{Results}
\label{sec:results}
	
\subsection{Offline Orientation Estimation Testing}
The results of the offline testing are shown in \fig{fig:offline_result}. The experimental results show that the angular velocity rectification model reduces the orientation estimation error by an average of 5.97\(^\circ\) on trained objects and by 3.09\(^\circ\) on novel objects, demonstrating the effectiveness and generalization capability of the angular velocity rectification model.

\subsection{Online Active Manipulation Testing}

\begin{table}[t]
\centering
\caption{Quantitative Results on Online Testing}
\label{table:online_result}
\begin{tabular}{|c|c|c|c|c|}
\hline

\multirow{2}{*}{Object} & \multicolumn{2}{c|}{RMSE ($^\circ$) $\downarrow$} & \multicolumn{2}{c|}{RMS Jerk (${^\circ/\text{s}^3}$) $\downarrow$} \\
\cline{2-5} 
 & Open-loop & Our Method & Open-loop & Our Method \\
\hline
A1 & \textbf{7.51 }& 9.90  & \textbf{6.94}  &  9.21  \\
A2 &  28.01  &  \textbf{11.15} & 21.51 &  \textbf{20.60}  \\
A3 &  27.03 &  \textbf{10.51}  &  65.09 &  \textbf{24.85} \\
B1 &  20.32 & \textbf{ 9.14}  &   26.88  &  \textbf{10.90} \\
B2 &  13.62 &  \textbf{10.83} & 30.65 & \textbf{24.07} \\
B3 &  22.62 &  \textbf{8.46}  &  29.77 & \textbf{24.91}  \\
C1 &  23.51 &  \textbf{10.39} &  40.32 & \textbf{37.91}  \\
C2 &  22.53 & \textbf{ 10.01} &  32.49 & \textbf{ 27.07} \\
C3 &  17.14 &  \textbf{9.11}  &  \textbf{39.09} & 39.48 \\
N1 &  25.58 &\textbf{ 18.41}  &  51.81&  \textbf{38.11} \\
N2 &  19.60 &\textbf{ 17.96}  &  \textbf{24.68} &  37.28 \\
N3 &  14.24 & \textbf{9.26}  &  20.21& \textbf {15.71 } \\
\hline
\end{tabular}
\end{table}

\fig{fig:online_result} shows the orientation trajectories of four representative objects using open-loop and our method. We evaluate the accuracy of trajectory control using root mean square error (RMSE) between the desired and actual trajectories, and the smoothness of trajectory using root mean square (RMS) jerk. The quantitative results for each objects under two methods are presented in \tab{table:online_result}. Our method achieved more precise trajectory control than open-loop method with RMSE being less than $12^\circ$ on nine trained objects and less than $19^\circ$ on three novel objects. Also, our method exhibits smoother trajectory control on seven out of the nine trained objects and two out of three novel objects. The results indicate that our method shows significant improvements in both the accuracy and smoothness of trajectory control, and its performance on novel objects further highlights its generalization potential.


\section{Conclusion and Future work}
\label{sec:conclusion}
	This paper presents \sensor{}, a novel vision-based tactile sensor with active surface. Our experimental results demonstrate that its design, which integrates tactile perception with active surface, significantly enhances the dexterity, robustness, and efficiency of in-hand object manipulation. 
Additionally, its smaller corner radius compared to rollers of the same size allows it to grasp smaller objects, broadening its applicability across various grippers. 
Future work includes making the sensor more compact, reducing the corner radius, and mounting the sensor on the fingers of multi-fingered dexterous hand. In terms of algorithms, we will investigate the application of imitation learning and reinforcement learning for controlling grippers equipped with this type of sensor.


\footnotesize{
\bibliographystyle{IEEEtran}
\bibliography{reference}

\begin{thebibliography}{10}
\providecommand{\url}[1]{#1}
\csname url@samestyle\endcsname
\providecommand{\newblock}{\relax}
\providecommand{\bibinfo}[2]{#2}
\providecommand{\BIBentrySTDinterwordspacing}{\spaceskip=0pt\relax}
\providecommand{\BIBentryALTinterwordstretchfactor}{4}
\providecommand{\BIBentryALTinterwordspacing}{\spaceskip=\fontdimen2\font plus
\BIBentryALTinterwordstretchfactor\fontdimen3\font minus
  \fontdimen4\font\relax}
\providecommand{\BIBforeignlanguage}[2]{{%
\expandafter\ifx\csname l@#1\endcsname\relax
\typeout{** WARNING: IEEEtran.bst: No hyphenation pattern has been}%
\typeout{** loaded for the language `#1'. Using the pattern for}%
\typeout{** the default language instead.}%
\else
\language=\csname l@#1\endcsname
\fi
#2}}
\providecommand{\BIBdecl}{\relax}
\BIBdecl

\bibitem{shaw2023leap}
K.~Shaw, A.~Agarwal, and D.~Pathak, ``Leap hand: Low-cost, efficient, and
  anthropomorphic hand for robot learning,'' \emph{arXiv preprint
  arXiv:2309.06440}, 2023.

\bibitem{wei2024wearable}
D.~Wei and H.~Xu, ``A wearable robotic hand for hand-over-hand imitation
  learning,'' in \emph{2024 IEEE International Conference on Robotics and
  Automation (ICRA)}.\hskip 1em plus 0.5em minus 0.4em\relax IEEE, 2024, pp.
  18\,113--18\,119.

\bibitem{si2024deltahands}
Z.~Si, K.~Zhang, O.~Kroemer, and F.~Z. Temel, ``Deltahands: A synergistic
  dexterous hand framework based on delta robots,'' \emph{IEEE Robotics and
  Automation Letters}, 2024.

\bibitem{yuan2017gelsight}
W.~Yuan, S.~Dong, and E.~H. Adelson, ``Gelsight: High-resolution robot tactile
  sensors for estimating geometry and force,'' \emph{Sensors}, vol.~17, no.~12,
  p. 2762, 2017.

\bibitem{qi2023general}
H.~Qi, B.~Yi, S.~Suresh, M.~Lambeta, Y.~Ma, R.~Calandra, and J.~Malik,
  ``General in-hand object rotation with vision and touch,'' in
  \emph{Conference on Robot Learning}.\hskip 1em plus 0.5em minus 0.4em\relax
  PMLR, 2023, pp. 2549--2564.

\bibitem{romero2024eyesight}
B.~Romero, H.-S. Fang, P.~Agrawal, and E.~Adelson, ``Eyesight hand: Design of a
  fully-actuated dexterous robot hand with integrated vision-based tactile
  sensors and compliant actuation,'' \emph{arXiv preprint arXiv:2408.06265},
  2024.

\bibitem{ma2024gellink}
Y.~Ma, J.~A. Zhao, and E.~Adelson, ``Gellink: A compact multi-phalanx finger
  with vision-based tactile sensing and proprioception,'' in \emph{2024 IEEE
  International Conference on Robotics and Automation (ICRA)}.\hskip 1em plus
  0.5em minus 0.4em\relax IEEE, 2024, pp. 1107--1113.

\bibitem{yuanDesignControlRoller2020}
S.~Yuan, L.~Shao, C.~L. Yako, A.~Gruebele, and J.~K. Salisbury, ``Design and
  {{Control}} of {{Roller Grasper V2}} for {{In-Hand Manipulation}},'' in
  \emph{2020 {{IEEE}}/{{RSJ International Conference}} on {{Intelligent
  Robots}} and {{Systems}} ({{IROS}})}, 2020, pp. 9151--9158.

\bibitem{ko2020tendon}
T.~Ko, ``A tendon-driven robot gripper with passively switchable underactuated
  surface and its physics simulation based parameter optimization,'' \emph{IEEE
  Robotics and Automation Letters}, vol.~5, no.~4, pp. 5002--5009, 2020.

\bibitem{li2023active}
S.~Li, F.~Wan, and C.~Song, ``Active surface with passive omni-directional
  adaptation of soft polyhedral fingers for in-hand manipulation,'' \emph{arXiv
  preprint arXiv:2311.14974}, 2023.

\bibitem{cai2023hand}
Y.~Cai and S.~Yuan, ``In-hand manipulation in power grasp: Design of an
  adaptive robot hand with active surfaces,'' in \emph{2023 IEEE International
  Conference on Robotics and Automation (ICRA)}.\hskip 1em plus 0.5em minus
  0.4em\relax IEEE, 2023, pp. 10\,296--10\,302.

\bibitem{xiang2024adaptive}
S.~Xiang, J.~Li, Y.~Zhang, Y.~Yang, J.~Liu, and Z.~Liu, ``Adaptive wrapping
  with active elastic band-based gripper for stable in-hand manipulation,''
  \emph{Sensors and Actuators A: Physical}, vol. 377, p. 115743, 2024.

\bibitem{jiang2024rotipbot}
J.~Jiang, X.~Zhang, D.~F. Gomes, T.-T. Do, and S.~Luo, ``Rotipbot: Robotic
  handling of thin and flexible objects using rotatable tactile sensors,''
  \emph{arXiv preprint arXiv:2406.09332}, 2024.

\bibitem{yuan2023tactile}
S.~Yuan, S.~Wang, R.~Patel, M.~Tippur, C.~Yako, E.~Adelson, and K.~Salisbury,
  ``Tactile-reactive roller grasper,'' \emph{arXiv preprint arXiv:2306.09946},
  2023.

\bibitem{lin2023dtact}
C.~Lin, Z.~Lin, S.~Wang, and H.~Xu, ``Dtact: A vision-based tactile sensor that
  measures high-resolution 3d geometry directly from darkness,'' in \emph{2023
  IEEE International Conference on Robotics and Automation (ICRA)}.\hskip 1em
  plus 0.5em minus 0.4em\relax IEEE, 2023, pp. 10\,359--10\,366.

\bibitem{yang2023tacgnn}
L.~Yang, B.~Huang, Q.~Li, Y.-Y. Tsai, W.~W. Lee, C.~Song, and J.~Pan, ``Tacgnn:
  Learning tactile-based in-hand manipulation with a blind robot using
  hierarchical graph neural network,'' \emph{IEEE Robotics and Automation
  Letters}, vol.~8, no.~6, pp. 3605--3612, 2023.

\bibitem{yin2023rotating}
Z.-H. Yin, B.~Huang, Y.~Qin, Q.~Chen, and X.~Wang, ``Rotating without seeing:
  Towards in-hand dexterity through touch,'' \emph{arXiv preprint
  arXiv:2303.10880}, 2023.

\bibitem{guzey2023dexterity}
I.~Guzey, B.~Evans, S.~Chintala, and L.~Pinto, ``Dexterity from touch:
  Self-supervised pre-training of tactile representations with robotic play,''
  \emph{arXiv preprint arXiv:2303.12076}, 2023.

\bibitem{zhao2023gelsight}
J.~Zhao and E.~H. Adelson, ``Gelsight svelte: A human finger-shaped
  single-camera tactile robot finger with large sensing coverage and
  proprioceptive sensing,'' in \emph{2023 IEEE/RSJ International Conference on
  Intelligent Robots and Systems (IROS)}.\hskip 1em plus 0.5em minus
  0.4em\relax IEEE, 2023, pp. 8979--8984.

\bibitem{van2020large}
L.~Van~Duong \emph{et~al.}, ``Large-scale vision-based tactile sensing for
  robot links: Design, modeling, and evaluation,'' \emph{IEEE Transactions on
  Robotics}, vol.~37, no.~2, pp. 390--403, 2020.

\bibitem{tincani2012velvet}
V.~Tincani, M.~G. Catalano, E.~Farnioli, M.~Garabini, G.~Grioli, G.~Fantoni,
  and A.~Bicchi, ``Velvet fingers: A dexterous gripper with active surfaces,''
  in \emph{2012 IEEE/RSJ International Conference on Intelligent Robots and
  Systems}.\hskip 1em plus 0.5em minus 0.4em\relax IEEE, 2012, pp. 1257--1263.

\bibitem{yuan2020design}
S.~Yuan, A.~D. Epps, J.~B. Nowak, and J.~K. Salisbury, ``Design of a
  roller-based dexterous hand for object grasping and within-hand
  manipulation,'' in \emph{2020 IEEE International Conference on Robotics and
  Automation (ICRA)}.\hskip 1em plus 0.5em minus 0.4em\relax IEEE, 2020, pp.
  8870--8876.

\bibitem{lepert2023hand}
M.~Lepert, C.~Pan, S.~Yuan, R.~Antonova, and J.~Bohg, ``In-hand manipulation of
  unknown objects with tactile sensing for insertion,'' in \emph{Embracing
  Contacts-Workshop at ICRA 2023}, 2023.

\bibitem{johnson2009retrographic}
M.~K. Johnson and E.~H. Adelson, ``Retrographic sensing for the measurement of
  surface texture and shape,'' in \emph{2009 IEEE Conference on Computer Vision
  and Pattern Recognition}.\hskip 1em plus 0.5em minus 0.4em\relax IEEE, 2009,
  pp. 1070--1077.

\bibitem{tippur2023gelsight360}
M.~H. Tippur and E.~H. Adelson, ``Gelsight360: An omnidirectional camera-based
  tactile sensor for dexterous robotic manipulation,'' in \emph{2023 IEEE
  International Conference on Soft Robotics (RoboSoft)}.\hskip 1em plus 0.5em
  minus 0.4em\relax IEEE, 2023, pp. 1--8.

\bibitem{lin20239dtact}
C.~Lin, H.~Zhang, J.~Xu, L.~Wu, and H.~Xu, ``9dtact: A compact vision-based
  tactile sensor for accurate 3d shape reconstruction and generalizable 6d
  force estimation,'' \emph{IEEE Robotics and Automation Letters}, 2023.

\bibitem{ma2021extrinsic}
D.~Ma, S.~Dong, and A.~Rodriguez, ``Extrinsic contact sensing with
  relative-motion tracking from distributed tactile measurements,'' in
  \emph{2021 IEEE international conference on robotics and automation
  (ICRA)}.\hskip 1em plus 0.5em minus 0.4em\relax IEEE, 2021, pp.
  11\,262--11\,268.

\bibitem{liu2023enhancing}
Y.~Liu, X.~Xu, W.~Chen, H.~Yuan, H.~Wang, J.~Xu, R.~Chen, and L.~Yi,
  ``Enhancing generalizable 6d pose tracking of an in-hand object with tactile
  sensing,'' \emph{IEEE Robotics and Automation Letters}, 2023.

\bibitem{wang2021gelsight}
S.~Wang, Y.~She, B.~Romero, and E.~Adelson, ``Gelsight wedge: Measuring
  high-resolution 3d contact geometry with a compact robot finger,'' in
  \emph{2021 IEEE International Conference on Robotics and Automation
  (ICRA)}.\hskip 1em plus 0.5em minus 0.4em\relax IEEE, 2021, pp. 6468--6475.

\bibitem{rusinkiewicz2001efficient}
S.~Rusinkiewicz and M.~Levoy, ``Efficient variants of the icp algorithm,'' in
  \emph{Proceedings third international conference on 3-D digital imaging and
  modeling}.\hskip 1em plus 0.5em minus 0.4em\relax IEEE, 2001, pp. 145--152.

\bibitem{garrido2014automatic}
S.~Garrido-Jurado, R.~Mu{\~n}oz-Salinas, F.~J. Madrid-Cuevas, and M.~J.
  Mar{\'\i}n-Jim{\'e}nez, ``Automatic generation and detection of highly
  reliable fiducial markers under occlusion,'' \emph{Pattern Recognition},
  vol.~47, no.~6, pp. 2280--2292, 2014.

\end{thebibliography}
}

\end{document}